%% file: main.tex
\documentclass{article}

\usepackage{microtype}
\usepackage{graphicx}
\usepackage{subfigure}
\usepackage{booktabs} % for professional tables
\usepackage{glossaries}

\usepackage{hyperref}

\usepackage[accepted]{icml2023-diffxyz}

\usepackage{amsmath}
\usepackage{amssymb}
\usepackage{mathtools}
\usepackage{amsthm}

\DeclareMathOperator{\E}{\mathbb{E}}

\usepackage[capitalize,noabbrev]{cleveref}

\theoremstyle{plain}

\theoremstyle{definition}

\theoremstyle{remark}

\usepackage[textsize=tiny]{todonotes}

\icmltitlerunning{Stochastic Gradient Bayesian Optimal Experimental Designs for Simulation-based Inference}

\begin{document}

\twocolumn[
\icmltitle{Stochastic Gradient Bayesian Optimal Experimental Designs \\
                for Simulation-based Inference}

% It is OKAY to include author information, even for blind
% submissions: the style file will automatically remove it for you
% unless you've provided the [accepted] option to the icml2023
% package.

% List of affiliations: The first argument should be a (short)
% identifier you will use later to specify author affiliations
% Academic affiliations should list Department, University, City, Region, Country
% Industry affiliations should list Company, City, Region, Country

% You can specify symbols, otherwise they are numbered in order.
% Ideally, you should not use this facility. Affiliations will be numbered
% in order of appearance and this is the preferred way.
\icmlsetsymbol{equal}{*}

\begin{icmlauthorlist}
\icmlauthor{Vincent D. Zaballa}{yyy}
\icmlauthor{Elliot E. Hui}{yyy}

%\icmlauthor{}{sch}
%\icmlauthor{}{sch}
\end{icmlauthorlist}

\icmlaffiliation{yyy}{Department of Biomedical Engineering, University of California, Irvine, Irvine, CA, USA}
% \icmlaffiliation{comp}{Company Name, Location, Country}
% \icmlaffiliation{sch}{School of ZZZ, Institute of WWW, Location, Country}

\icmlcorrespondingauthor{Vincent Zaballa}{vzaballa@uci.edu}
\icmlcorrespondingauthor{Elliot Hui}{eehui@uci.edu}

% You may provide any keywords that you
% find helpful for describing your paper; these are used to populate
% the "keywords" metadata in the PDF but will not be shown in the document
\icmlkeywords{Machine Learning, ICML}

\vskip 0.3in
]

% this must go after the closing bracket ] following \twocolumn[ ...

% This command actually creates the footnote in the first column
% listing the affiliations and the copyright notice.
% The command takes one argument, which is text to display at the start of the footnote.
% The \icmlEqualContribution command is standard text for equal contribution.
% Remove it (just {}) if you do not need this facility.

\printAffiliationsAndNotice{}  % leave blank if no need to mention equal contribution
% \printAffiliationsAndNotice{\icmlEqualContribution} % otherwise use the standard text.

\begin{abstract}
Simulation-based inference (SBI) methods tackle complex scientific models with challenging inverse problems. However, SBI models often face a significant hurdle due to their non-differentiable nature, which hampers the use of gradient-based optimization techniques. Bayesian Optimal Experimental Design (BOED) is a powerful approach that aims to make the most efficient use of experimental resources for improved inferences. While stochastic gradient BOED methods have shown promising results in high-dimensional design problems, they have mostly neglected the integration of BOED with SBI due to the difficult non-differentiable property of many SBI simulators. In this work, we establish a crucial connection between ratio-based SBI inference algorithms and stochastic gradient-based variational inference by leveraging mutual information bounds. This connection allows us to extend BOED to SBI applications, enabling the simultaneous optimization of experimental designs and amortized inference functions. We demonstrate our approach on a simple linear model and offer implementation details for practitioners.
\end{abstract}

\glsresetall
\input{introduction}
\input{background}

\input{method}
\input{experiments}

\input{discussion}

\section*{Acknowledgements}

This research was funded by the National Institute of General Medical Sciences (NIGMS) of the National Institutes of Health (NIH) under award number 1F31GM145188-01. We would like to thank Adam Foster and members of the Elowitz Lab for helpful discussions.

% In the unusual situation where you want a paper to appear in the
% references without citing it in the main text, use \nocite

% \bibliography{example_paper}
% \bibliographystyle{icml2023}
\bibliographystyle{icml2023} % Choose a bibliography style
\bibliography{references}

%%%%%%%%%%%%%%%%%%%%%%%%%%%%%%%%%%%%%%%%%%%%%%%%%%%%%%%%%%%%%%%%%%%%%%%%%%%%%%%
%%%%%%%%%%%%%%%%%%%%%%%%%%%%%%%%%%%%%%%%%%%%%%%%%%%%%%%%%%%%%%%%%%%%%%%%%%%%%%%
% APPENDIX
%%%%%%%%%%%%%%%%%%%%%%%%%%%%%%%%%%%%%%%%%%%%%%%%%%%%%%%%%%%%%%%%%%%%%%%%%%%%%%%
%%%%%%%%%%%%%%%%%%%%%%%%%%%%%%%%%%%%%%%%%%%%%%%%%%%%%%%%%%%%%%%%%%%%%%%%%%%%%%%
\newpage
\appendix
\onecolumn

\section{Design Gradients of LF-PCE}

For LF-PCE, we need unbiased gradient estimators of the information bounds. A normalizing flow can be seen as a reparameterized distribution, which allows for calculating the gradient with respect to designs $\nabla_\xi \boldsymbol{f}^{-1} (u; \theta, \xi)$. In practice, since we are evaluating the log probability of a data point, we would actually evaluate the inverse direction of a flow $\nabla_\xi \boldsymbol{f} (y; \theta, \xi) $ at the base distribution $p_u(\bold{u})$, which is usually a Gaussian distribution and evaluated by maximum likelihood.

More formally, following equation \ref{eq:snl_loss}, the gradient with respect to $\xi$ is
\begin{equation}
    \nabla_\xi \mathcal{L}(\xi) \approx - \frac{1}{N} \nabla_\xi \sum_n \log p_u( \boldsymbol{f}^{-1}(\bold{y}_n;\phi, \theta, \xi) + \log \rvert \det \boldsymbol{J}(\boldsymbol{f}^{-1})(\bold{y}_n;\phi, \theta, \xi) \rvert ),
\end{equation}
which is tractable as long as we can compute $\boldsymbol{f}^{-1}$, its Jacobian determinant, and evaluate the base density, $p_u(u)$, which is tractable for a base Gaussian distribution. Given this gradient, we can plug this into the gradient of LF-PCE to estimate the gradient of the information bound:
\begin{equation}
	\frac{\partial{I}_{LF-PCE}}{\partial\xi} = \E_{p(\theta_0)p(y|\theta,\xi)q(\theta_{1:L}|y)} \left[ \frac{\partial g}{\partial\xi} + g \cdot \frac{\partial}{\partial \xi}\log p\phi(y|\theta_0,\xi) \right],
	\label{eq:grad_lf_pce}
\end{equation}
where
\begin{equation}
	g(y, \theta_{0:L}, \phi, \xi) = \log \frac{p_\phi(y|\theta_0,\xi)}{\frac{1}{L+1}\sum_{\ell=0}^L p_\phi(y|\theta_\ell,\xi)}.
\end{equation}

\section{Evaluation of Linear Model Designs and Posteriors}

\begin{figure*}[!htb]
  \centering
  \includegraphics[width=\textwidth]{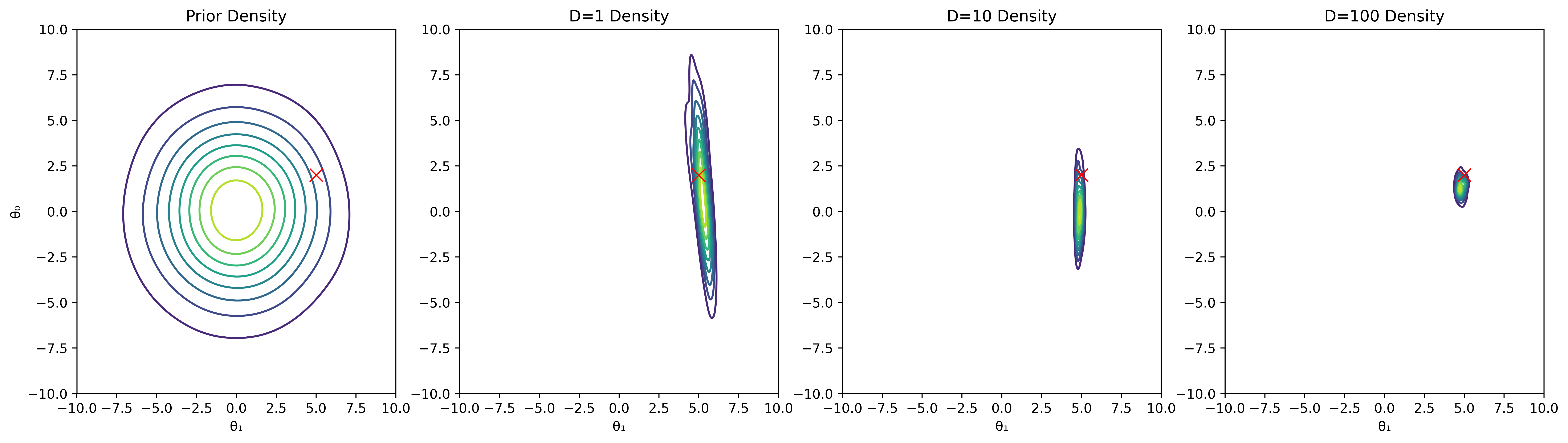}
  \vspace{-20pt}
  \caption{Comparison of the prior density the posterior achieved by the different design dimensional normalizing flows evaluated at an optimal design $p(\theta|y_o, \xi^*) \propto p_\phi (y_o | \theta, \xi^*)p(\theta)$. The red cross denotes the true model parameters. }
  % \vspace{-15pt}}
  \label{fig:posteriors}
\end{figure*}

We evaluated the efficacy of the neural density estimator trained using the LF-PCE loss function to infer a held out true parameter value in Figure \ref{fig:posteriors} by MCMC. We provide a quantitative evaluation of the posteriors in Table \ref{tab:posts_table}. The posteriors can be improved by computationally efficient methods such as Sampling Importance Resampling, or used in SBI algorithms that use sequential methods to refine the neural density estimator.

\section{Evaluation of Posterior Predictive Distribution}

\begin{figure*}[!ht]
  \centering
  \vspace{-140pt}
  \includegraphics[scale=0.5]{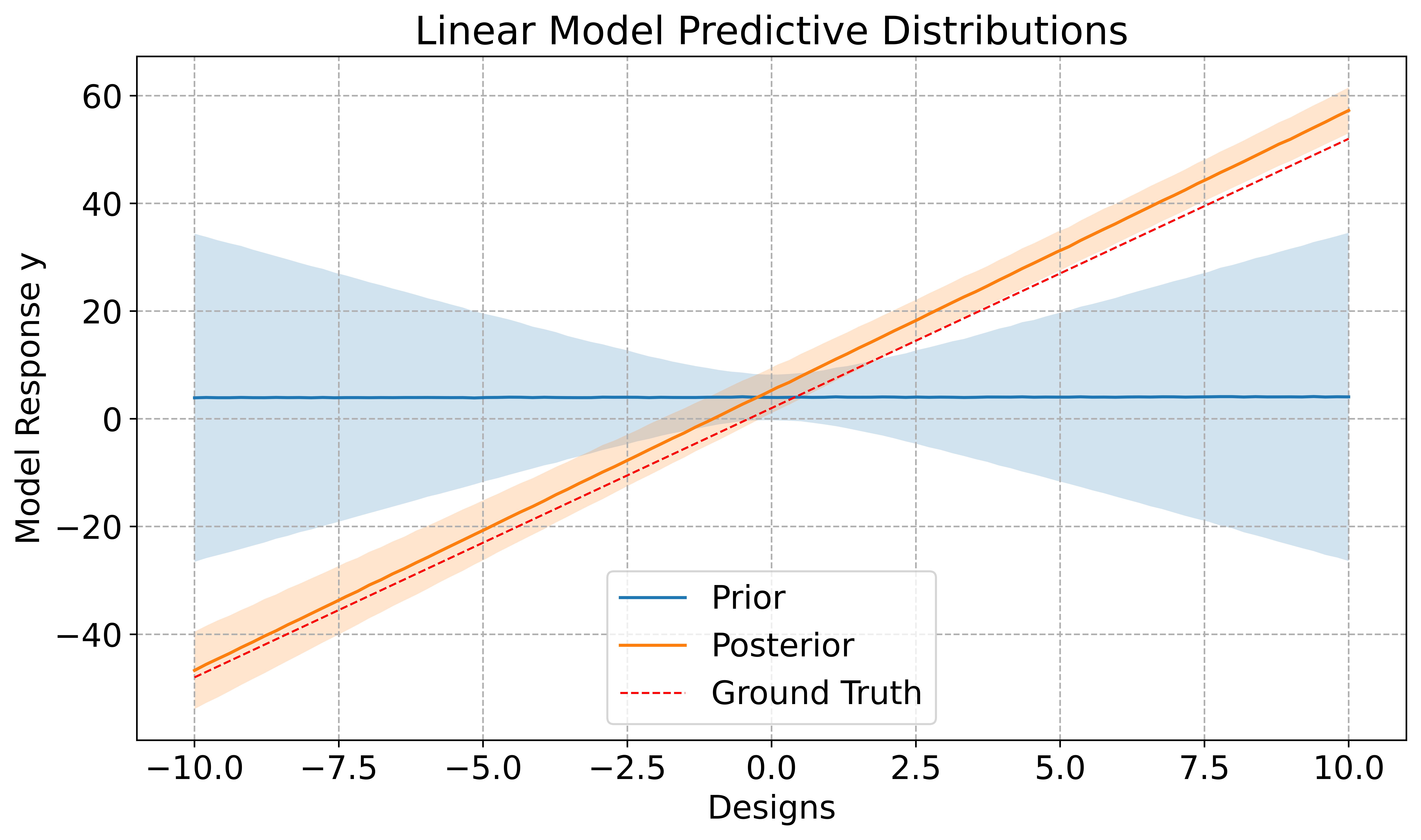}
  \caption{Prior predictive (blue) and posterior predictive (orange) distributions with the ground truth liner model (dotted red) for the single design case where D=1.  }
  % \vspace{-15pt}}
  \label{fig:posterior_predictive}
\end{figure*}

As a reference, we plot the prior and posterior predictive plots for the 1-dimensional optimal design in Figure \ref{fig:posterior_predictive}. An insight into the optimal experimental design problem is that the designs closer to where the prior distribution has more noise will lead to more clarification in a performed experiment, which is why the most optimal designs will be at the boundaries for the noisy linear model.

%%%%%%%%%%%%%%%%%%%%%%%%%%%%%%%%%%%%%%%%%%%%%%%%%%%%%%%%%%%%%%%%%%%%%%%%%%%%%%%
%%%%%%%%%%%%%%%%%%%%%%%%%%%%%%%%%%%%%%%%%%%%%%%%%%%%%%%%%%%%%%%%%%%%%%%%%%%%%%%

\end{document}

%% file: introduction.tex
% !TEX root =  main.tex

\section{Introduction}
\label{sec:introduction}

Many scientific models are defined by a simulator that defines an output $y$ determined by the inputs, or designs, to a system, $\xi$, and parameters that define how the model transforms the inputs to outputs, $\theta$. Inferring a distribution of parameters given data $p(\theta | y, \xi)$ is of central importance in Bayesian statistics and can be seen as a form of solving the inverse problem for a given simulator\cite{lindley1972}. In SBI, a simulator forms an implicit probability distribution known as the likelihood $p(y|\theta, \xi)$ that is used with the prior of the model parameters $p(\theta)$ to estimate the posterior probability of the model parameters given the observed data, $p(\theta | y, \xi)$ \cite{Cranmer2020}. Recent SBI methods have use deep learning-based models to infer either the intractable likelihood or posterior using density estimators for both, or classifiers to estimate the likelihood-to-evidence ratio, $\frac{p(\theta|y, \xi)}{p(\theta| \xi)} = \frac{p(y |\theta, \xi)}{p(y|\xi)} = \frac{p(y, \theta| \xi)}{p(\theta)p(y|\xi)}$. 

However, inferring the likelihood, posterior, or ratio is a computationally expensive process that depends on observed data $y_o$, to compute. Recent work questioned the validity of this expensive computational process used in SBI if using the wrong simulator for the true data generating process \cite{Cannon2022}. Naïve conclusions can be made if using the wrong model of the underlying scientific phenomenon, or the model is not close enough to the real data generating process, which motivates the use of optimal experimental designs in SBI methods.

Bayesian optimal experimental design (BOED) has shown promise as a way to optimize experiments given a model, the simulator, and priors of the parameters of interest. BOED works by determining the information gain of a proposed experimental design, $\xi$, on the parameters of the model of interest
\begin{equation}
    IG(y, \xi) = H[p(\theta)] - H[p(\theta|y, \xi)].
\end{equation}
The information gain can only be evaluated after an experiment but another quantity, the Expected Information Gain (EIG), $I(\xi)$, can be used as a proxy for the information gained in an experiment 
\begin{equation}
    I(\xi) \triangleq \E_{p(y|\xi)}\left[ H[p(\theta)] - H[p(\theta|y,\xi)] \right],
\end{equation}
The intuition behind this process is we must ask ourselves, which experimental design and outcome would be most surprising given what we assume, or know, about the system when conducting the experiment. This can be rewritten into the form of calculating the mutual information between the observed data and unknown parameters 
\begin{equation}
    I(\xi) = \text{MI}_\xi(\theta;y) = \E_{p(\theta)p(y|\theta,\xi)}\left[\log \frac{p(y|\theta,\xi)}{p(y|\xi)} \right].
    \label{eq:infoGain}
\end{equation}
Early BOED work focused on estimating the mutual information then using that estimate as the surrogate function in an outer optimizer, such as Bayesian optimization \cite{Foster2019a, kleinegesse2019efficient}. This double loop of optimization was inefficient and lead to development of methods to simultaneously optimize the design and mutual information in a single optimization process. However, this unified optimization depended on an unnormalized likelihood and posterior approximation \cite{Foster2019} or an implicit likelihood with a simulator that has a differentiable functional form \cite{Kleinegesse2021}. 

We present a method to simultaneously optimize designs and the mutual information for the remaining set of models, implicit likelihoods without a differentiable simulator, which are typically used in the SBI literature. We additionally make a link to how we can use a generative model in Contrastive Precitive Coding. We show:
\begin{itemize}
    \item A differentiable objective for simultaneously optimizing the mutual information and likelihood for SBI-based models.
    \item A connection between Likelihood-Free based methods for BOED and contrastive ratio estimation (CRE) methods for SBI models.
    \item Experimental validation of the unified objective on a simple linear model.
\end{itemize}

%%% Local Variables:
%%% mode: latex
%%% TeX-master: "../main"
%%% End:

%% file: background.tex
% !TEX root =  main.tex

\section{Background}
\label{sec:background}

Previous work in SBI methods have focused on improving methods based on given, observed, data $y_o$, \cite{Papamakarios2016,Papamakarios2018,Durkan2020,greenberg2019automatic} whereas BOED has focused on determining an optimal design $\xi^*$, based on various bounds of MI between $y$ and $\theta$. While these aims seem to be unrelated, we will show how they can be performed simultaneously for SBI methods that rely on potentially stochastic simulators that act as black-box functions.

\subsection{Simulation-Based Inference}

In many scientific disciplines, it is desirable to infer a distribution of parameters $\theta$, of a potentially stochastic model, or simulator, given observations, $y_o$. The closed-box simulator may depend on random numbers $z$, such as in stochastic differential equations, and previous experimental designs $\xi$, such that the simulator takes the form $y = g(\theta, \xi, z)$. When a likelihood is not available, Approximate Bayesian Computation (ABC) methods can be used, \cite{sisson2018overview} which aim to infer the likelihood of parameters of the simulator that are within an $\epsilon$ ball, $B_\epsilon(y)$, of the observed data $y\coloneqq y_o$, resulting in the likelihood $p(\|y - y_o\| < \epsilon | \theta)$. However, recent SBI methods have outperform ABC methods in inference tasks \cite{lueckmann2021benchmarking}. By using a simulator to simulate the joint data distribution  $(\theta, y) \sim p(y | \theta)$, drawn from a prior $\theta \sim p(\theta)$, we can obtain an amortized likelihood $p_\phi(y | \theta)$ or posterior $p_\phi(\theta | y)$ by training a neural density estimator, such as a normalizing flow, with parameters $\phi$, or estimate the likelihood-to-evidence ratio $\exp f_\phi (\theta, y) \approx \frac{p(y|\theta)}{p(y)}$, by training a classifier to distinguish parameters used to simulate an observed values, $y$. Different SBI methods can be used in inference for downstream applications depending on the desiderata of the inference task. For example, one might use an amortized posterior approximation if there are many different data samples to evaluate, whereas an ensemble of ratios \cite{hermanscrisis} has been shown to perform more robustly on Simulation-Based Calibration (SBC) tests \cite{Talts2018} at the cost of increased computational complexity.

There are many SBI methods proposed for approximating the likelihood, posterior, or ratio. We review the relevant ones to our method here. See \cite{lueckmann2021benchmarking} for a more thorough review and benchmark of SBI methods.

\textbf{Neural Likelihood Estimation}
We can use data from the joint distribution to train a conditional neural density-based likelihood function. If we take a dataset of samples $\{ y_n, \theta_n \}_{1:N}$ obtained from a simulator as previously described, we can train a conditional density estimator $p_\phi (y|\theta)$ to model the likelihood by maximizing the total log likelihood of $\sum_n \log p_\phi (y_n|\theta_n)$, which is approximately equivalent to minimizing the loss
\begin{equation}
    \mathcal{L}(\phi) = \E_{p(\boldsymbol{\theta})}(D_{\text{KL}} (p(y | \theta ) \| p_\phi (y | \theta) ) + \text{const},
    \label{eq:snl_loss}
\end{equation}
where the Kullback-Leibler divergence is minimized when $p_\phi (y|\theta)$ approaches $p (y | \theta)$. SBI methods would then condition this likelihood on observed data, $y_o$, and refine the likelihood estimate by resetting the prior to become the new posterior samples via Markov Chain Monte Carlo (MCMC) sampling of the approximate likelihood $p(\theta) \coloneqq p(\theta | y_o) \propto p_\phi (y_o|\theta) p(\theta) $ and training a new neural density estimator of the likelihood \cite{Papamakarios2018, lueckmann2018likelihood}. This is Sequential Neural Likelihood (SNL) which we forego as we focus on the preliminary step of optimizing an experimental design without $y_o$.

\subsection{Bayesian Optimal Experimental Design}

Following from equation \ref{eq:infoGain}, \cite{Foster2019} proposed the prior contrastive estimation (PCE) lower bound of the MI
\begin{equation}
    {I}_{PCE}(\xi, L) \triangleq \E \left[\log \frac{p(y|\theta_0,\xi)}{\frac{1}{L+1}\sum_{\ell=0}^{L} p(y|\theta_\ell,\xi)}\right],
    \label{eq:pce}
\end{equation} 
where the expectation is over $p(\theta_0)p(y| \theta_0, \xi)p(\theta_{1:L})$ and $\xi$ is the proposed design, $\theta_0$ is the original parameter that generated data $y$, and $L$ is the number of contrastive samples. The PCE bound is appropriate in BOED when the prior and posterior are similar enough that $p(\theta)$ is a suitable proposal distribution for $p(y|\xi)$. This bound has low variance but is upper-bounded by $\log L$, potentially leading to large bias but still demonstrated adequate performance on various benchmarks. Unfortunately, this bound requires a tractable likelihood function, which is not available in SBI applications.

%%% Local Variables:
%%% mode: latex
%%% TeX-master: "../main"
%%% End:

%% file: method.tex
% !TEX root =  main.tex

\section{SBI-based BOED}
\label{sec:method}
\subsection{Likelihood Free PCE}

We take inspiration from previous SBI and BOED methods to allow optimization of designs with respect to closed-box simulators that are modeled using normalizing flows. We start by noting how the loss function of contrastive ratio estimation (CRE) \cite{Durkan2020} lower bounds PCE 
\begin{align*}
\log \frac{\exp( f_\phi (\theta, y) )}{ \sum_{\ell=0}^L \exp ( f_\phi (\theta_{\ell}, y))} 
&\leq \log \frac{\exp( f_\phi (\theta, y) )}{ \frac{1}{1 + L} \sum_{\ell=0}^L \exp ( f_\phi (\theta_{\ell}, y))} \\
&= \log \frac{ p_\phi (y | \theta_0, \xi)}{\frac{1}{1 + L}\sum^L_{\ell = 0} p_\phi (y | \theta_l, \xi)},
\end{align*}
where $L$ is the number of contrastive samples and $f_\phi$ is a discriminative classifier, which holds for a single batch of data and constant experimental design, i.e. when $\xi$ is constant. We exchange an explicit likelihood in PCE with a neural density estimator to create Likelihood-Free PCE (LF-PCE). We now have a MI lower bound
\begin{equation}
    I(\xi, \phi, L) \geq \E \Bigg[\log \frac{ p_{\phi}(y | \theta_0, \xi)}{\frac{1}{1 + L}\sum^L_{\ell = 0} p_{\phi}(y | \theta_l, \xi)}  \Bigg],
\end{equation}
where the expectation is over $p(\theta_0)p(y|\theta_0, \xi)p(\theta_{1:L})$. We now can simultaneously optimizes designs and parameters of a neural density estimator. If we are to use a normalizing flow as $\exp f_\phi(y, \theta, \xi) = p_\phi(y|\theta, \xi)$, then the PCE lower bound of the MI holds since the distribution is normalized as normalizing flows are bounded functions \cite{Papamakarios2019}. We note that this can be an unstable objective as the data distribution of the flow will change as experimental designs change. However, the result is that it returns an amortized likelihood that can be evaluated on observed experimental data to return a posterior density or used in downstream inference algorithms, such as SNL. Finally, using a normalizing flow allows us to take gradients with respect to designs $\xi$, which we derive in Appendix A.

\textbf{Practical implementation of LF-PCE loss}
For LF-PCE training, stability of the density estimator is a challenge when optimizing the MI lower bound. To address this, we added a regularization term, $\lambda$, to both loss functions to help stabilize the training of the density estimator during design optimization
\begin{equation}
    \E \Bigg[\log \frac{ p_{\phi}(y | \theta_0, \xi)}{\frac{1}{1 + L}\sum^L_{\ell = 0} p_{\phi}(y | \theta_l, \xi)} + \lambda \cdot \log p_{\phi}(y | \theta_0, \xi)  \Bigg],
    \label{eq:lamb_lf_pce}
\end{equation}
where the expectation is over $p(\theta_0)p(y|\theta_0, \xi)p(\theta_{1:L})$.

\subsection{Connection to Generative MI Estimation}

The mutual information bound proposed by \cite{Foster2019} for PCE is similar to Contrastive Predictive Coding (CPC) \cite{poole2019variational, oord2018representation}, but where a generative model replaces a discriminative one and the random variable X corresponds to observed data and random variable Y to the prior distribution. In our formulation the bound of the MI depends on both the amount of training $tr \rightarrow \infty$ and number of contrastive samples $L \rightarrow \infty$ to approach the true MI. The generative approach to CPC can be simplified as 
\begin{equation}
    I_{PCE}(\phi) \coloneqq \E_P [ \log p_\phi (x|y) - \log p_\phi(x) ],
    \label{eq:MI_lf-pce}
\end{equation}
where $P$ is a random variable representing the joint distribution we obtain from our simulators $(x, y) \sim p(x|y)p(y)$ and $p_\phi(x)$ implicitly depends on the number of contrastive samples $L$ to approximate the marginal likelihood.

%%% Local Variables:
%%% mode: latex
%%% TeX-master: "../main"
%%% End:

%% file: experiments.tex
% !TEX root =  main.tex

\section{Experimental Evaluation}
\label{sec:experiments}

\subsection{Noisy Linear Model}

\begin{figure*}[!htb]
  \centering
  \includegraphics[width=\textwidth]{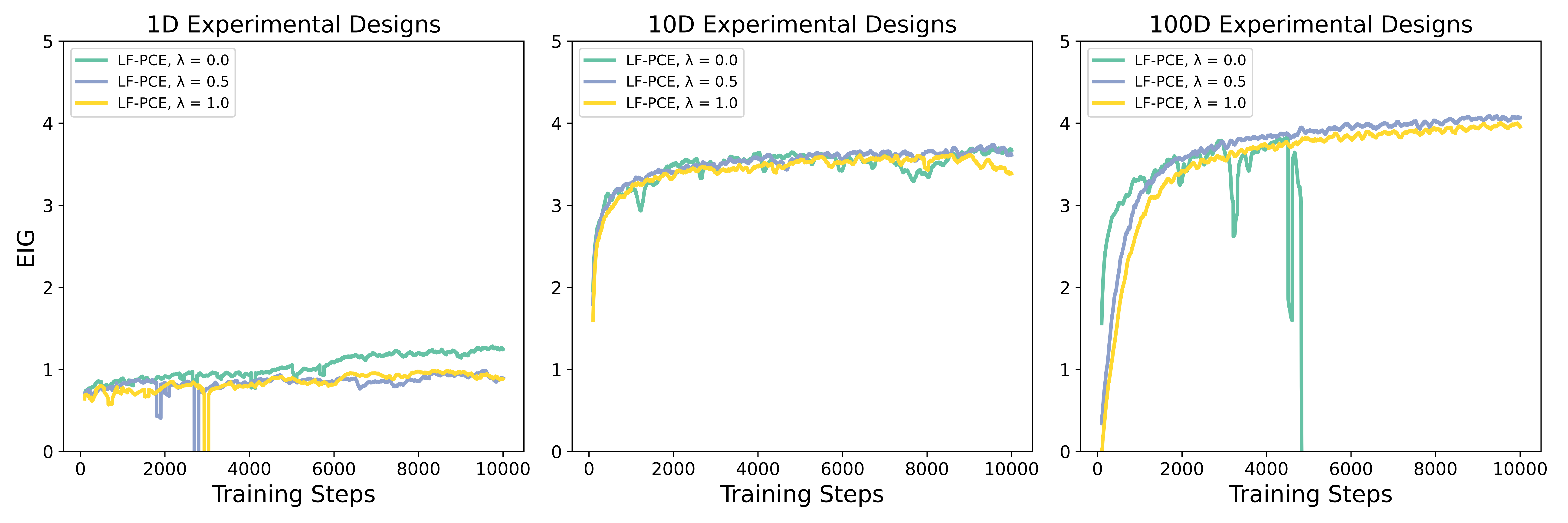}
  \vspace{-20pt}
  \caption{Comparison of the EIG across design dimensions, type of BOED, and $\lambda$ regularization for the noisy linear model examining the moving average over N=10 samples. For the single design dimension, LF-PCE with no $\lambda$ regularization outperforms in estimating a lower bound of the MI, which can translate to more informative experimental designs. In the higher-dimension design cases, LF-PCE increases its EIG with more designs, which is expected, but sees diminishing returns when expanding from 10D to 100D design evaluations. In the 100-dimensional design case, we see the benefit of using $\lambda$ regularization to stabilize the training of a neural density estimator in high-dimensional input space at the cost of slightly lower EIG.}
  % \vspace{-15pt}}
  \label{fig:linear_eig}
\end{figure*}

We follow \cite{Kleinegesse2020b} and evaluate optimal designs on a classic noisy linear model where a response variable $y$ has a linear relationship with experimental designs $\xi$, which is determined by values of the model parameters $\boldsymbol{\theta} = [\theta_0, \theta_1]$, which model the offset and gradient. We would like to optimize the value of $D$ measurements to estimate the posterior of $\boldsymbol{\theta}$, and so create a design vector $\boldsymbol{\xi} = [\xi_1, \dots, \xi_D]^{\mathsf{T}}$.  Each design, $\xi_i$ returns a measurement $y_i$, which results in the data vector $\bold{y} = [y_1, \dots, y_D]^{\mathsf{T}}$. We assume non-Gaussian noise sources, otherwise evaluating the posterior and MI would be trivial. We use a Gaussian noise source $\mathcal{N}(\epsilon; 0, 1)$ and Gamma noise source $\Gamma (\nu; 2, 2)$. The model is then 
\begin{equation}
    y = \theta_0\boldsymbol{1} + \theta_1 * \boldsymbol{\xi} + \boldsymbol{\epsilon} + \boldsymbol{\nu},
\end{equation}
where $\boldsymbol{\epsilon} = [\epsilon_1, \dots, \epsilon_D]^{\mathsf{T}}$ and $\boldsymbol{\nu} = [\nu_1, \dots, \nu_D]^{\mathsf{T}}$ are i.i.d. samples.  We evaluate LF-PCE on this model and examine how changing the $\lambda$ regularization parameter in \eqref{eq:lamb_lf_pce} influences the resulting mutual information bound and design quality for both models.

For each design dimension, D, we randomly initialize designs $\xi \in [-10, 10]$. For LF-PCE, we chose $N=10$, the number of non-contrastive samples $y \sim p(y|\xi, \theta_0)$, and $M=50$ contrastive samples for all experiments. For the neural spline flow, we chose 5 bijector layers, each with 4 bins, and 4 resnet multilayer perceptrons, each with 128 dimensions, for the neural network-based conditional networks. For both the neural density estimator's parameters $\phi$, and the designs $\xi$, we use the Adam optimizer \cite{kingma2014adam} with $\beta_1 = 0.9$ and $\beta_2 = 0.99$, with learning rate $\alpha=1e^{-3}$ for the neural density estimator and $\alpha=1e^{-2}$ for design optimization.

Examining the graph of the mutual information in Figure \ref{fig:linear_eig}, we see that LF-PCE lower bound steadily increases for all values of lambda; however, the stability of the optimization of the generative model's parameters diverges in higher design dimensions whenever $\lambda=0$. We see a general trend between exploration and exploitation in changing values of $\lambda$, where higher $\lambda$ values lead to lower MI lower-bound estimates and potentially more homogenous designs.

Using LF-PCE we obtain an amortized neural density estimator of the likelihood that is able to perform inference on observed data evaluated at the optimal design. For example, $p(\theta|y_o, \xi^*) \propto p_\phi (y_o | \theta, \xi^*)p(\theta)$ by MCMC sampling. We evaluate the posterior densities after optimizing on the LF-PCE lower bound in Appedix B and can see the mean and interquartile range in Table \ref{tab:posts_table}. We note that we were able to arrive at accurate and precise posterior estimates using the neural density estimator that simultaneously optimized an optimal design $\xi^*$, without any post-processing such as using SNL or Sampling Importance Resampling.

\begin{table}[!htb]
\centering
\small
\begin{tabular}{lcc}
\toprule
Design Dimension & $\theta_0$ & $\theta_1$ \\
\midrule
D=1   & $1.29 \pm 2.98$ & $5.20 \pm 0.41$ \\
D=10     & $0.07 \pm 1.40$ & $4.87 \pm 0.16$ \\
D=100      & $1.35 \pm 0.52$ & $4.81 \pm 0.20$ \\
\bottomrule
\end{tabular}
\caption{Posterior estimates mean and 68\% interquartile range after observing $\xi^*$ values for each design dimension only using the amortized likelihood approximation provided by the neural density estimator used in the LF-PCE training. The held-out parameter values that were used to generate $y_o$ were $\boldsymbol{\theta}_{\text{true}} = [2, 5]$. More design dimensions approach the true held-out parameter with increasing precision.
\vspace{-15pt}}
\label{tab:posts_table}
\end{table}

%%% Local Variables:
%%% mode: latex
%%% TeX-master: "../main"
%%% End:

%% file: discussion.tex
% !TEX root =  main.tex

\section{Discussion}
\label{sec:discussion}

We demonstrated a novel information bounds, $I_{LF-PCE}$, to perform gradient-based BOED using black-box simulators present in many SBI applications and obtained lower bounds of the EIG on a toy model across a range of experimental design dimensions to showcase its scalability. Optimizing designs in SBI applications provides a valuable preconditioning step to typical sequential SBI methods such as SNL that are based on observed experimental designs. Sidestepping Bayesian optimization can also help to accelerate model testing and feedback from real-world data. Future work will examine the tradeoff between design diversity for improved entropy reduction and neural density estimator robustness, similar to the exploration and exploitation tradeoff present in Bayesian optimization.

%%% Local Variables:
%%% mode: latex
%%% TeX-master: "../main"
%%% End:

%% file: main.bbl
\begin{thebibliography}{21}
\providecommand{\natexlab}[1]{#1}
\providecommand{\url}[1]{\texttt{#1}}
\expandafter\ifx\csname urlstyle\endcsname\relax
  \providecommand{\doi}[1]{doi: #1}\else
  \providecommand{\doi}{doi: \begingroup \urlstyle{rm}\Url}\fi

\bibitem[Cannon et~al.()Cannon, Ward, and Schmon]{Cannon2022}
Cannon, P., Ward, D., and Schmon, S.~M.
\newblock Investigating the impact of model misspecification in neural
  simulation-based inference.
\newblock \doi{10.48550/arxiv.2209.01845}.
\newblock URL \url{https://arxiv.org/abs/2209.01845v1}.

\bibitem[Cranmer et~al.(2020)Cranmer, Brehmer, and Louppe]{Cranmer2020}
Cranmer, K., Brehmer, J., and Louppe, G.
\newblock The frontier of simulation-based inference.
\newblock \emph{Proceedings of the National Academy of Sciences}, 117\penalty0
  (48):\penalty0 30055--30062, November 2020.
\newblock ISSN 0027-8424.
\newblock \doi{10.1073/pnas.1912789117}.
\newblock URL \url{http://arxiv.org/abs/1911.01429}.
\newblock arXiv: 1911.01429 Publisher: Proceedings of the National Academy of
  Sciences.

\bibitem[Durkan et~al.(2020)Durkan, Murray, and Papamakarios]{Durkan2020}
Durkan, C., Murray, I., and Papamakarios, G.
\newblock On contrastive learning for likelihood-free inference, February 2020.
\newblock URL \url{http://arxiv.org/abs/2002.03712}.
\newblock arXiv: 2002.03712 Publication Title: arXiv.

\bibitem[Foster et~al.(2019{\natexlab{a}})Foster, Jankowiak, Bingham, Horsfall,
  Teh, Rainforth, and Goodman]{Foster2019}
Foster, A., Jankowiak, M., Bingham, E., Horsfall, P., Teh, Y.~W., Rainforth,
  T., and Goodman, N.
\newblock Variational {Bayesian} optimal experimental design.
\newblock \emph{arXiv}, March 2019{\natexlab{a}}.
\newblock ISSN 23318422.
\newblock URL \url{http://arxiv.org/abs/1903.05480}.
\newblock arXiv: 1903.05480 Publisher: arXiv.

\bibitem[Foster et~al.(2019{\natexlab{b}})Foster, Jankowiak, O'Meara, Teh, and
  Rainforth]{Foster2019a}
Foster, A., Jankowiak, M., O'Meara, M., Teh, Y.~W., and Rainforth, T.
\newblock A unified stochastic gradient approach to designing
  {Bayesian}-optimal experiments.
\newblock \emph{arXiv}, November 2019{\natexlab{b}}.
\newblock ISSN 23318422.
\newblock URL \url{http://arxiv.org/abs/1911.00294}.
\newblock arXiv: 1911.00294 Publisher: arXiv.

\bibitem[Greenberg et~al.(2019)Greenberg, Nonnenmacher, and
  Macke]{greenberg2019automatic}
Greenberg, D., Nonnenmacher, M., and Macke, J.
\newblock Automatic posterior transformation for likelihood-free inference.
\newblock In \emph{International Conference on Machine Learning}, pp.\
  2404--2414. PMLR, 2019.

\bibitem[Hermans et~al.()Hermans, Delaunoy, Rozet, Wehenkel, Begy, and
  Louppe]{hermanscrisis}
Hermans, J., Delaunoy, A., Rozet, F., Wehenkel, A., Begy, V., and Louppe, G.
\newblock A crisis in simulation-based inference? beware, your posterior
  approximations can be unfaithful.
\newblock \emph{Transactions on Machine Learning Research}.

\bibitem[Kingma \& Ba(2014)Kingma and Ba]{kingma2014adam}
Kingma, D.~P. and Ba, J.
\newblock Adam: A method for stochastic optimization.
\newblock \emph{arXiv preprint arXiv:1412.6980}, 2014.

\bibitem[Kleinegesse \& Gutmann(2019)Kleinegesse and
  Gutmann]{kleinegesse2019efficient}
Kleinegesse, S. and Gutmann, M.~U.
\newblock Efficient bayesian experimental design for implicit models.
\newblock In \emph{The 22nd International Conference on Artificial Intelligence
  and Statistics}, pp.\  476--485. PMLR, 2019.

\bibitem[Kleinegesse \& Gutmann(2020)Kleinegesse and Gutmann]{Kleinegesse2020b}
Kleinegesse, S. and Gutmann, M.~U.
\newblock Bayesian experimental design for implicit models by mutual
  information neural estimation, February 2020.
\newblock URL \url{http://arxiv.org/abs/2002.08129}.
\newblock arXiv: 2002.08129 Publication Title: arXiv.

\bibitem[Kleinegesse \& Gutmann(2021)Kleinegesse and Gutmann]{Kleinegesse2021}
Kleinegesse, S. and Gutmann, M.~U.
\newblock Gradient-based {Bayesian} {Experimental} {Design} for {Implicit}
  {Models} using {Mutual} {Information} {Lower} {Bounds}.
\newblock May 2021.
\newblock URL \url{https://arxiv.org/abs/2105.04379v1}.
\newblock arXiv: 2105.04379.

\bibitem[Lindley(1972)]{lindley1972}
Lindley, D.~V.
\newblock \emph{Bayesian statistics, a review}, volume~2.
\newblock SIAM, 1972.

\bibitem[Lueckmann et~al.(2018)Lueckmann, Bassetto, Karaletsos, and
  Macke]{lueckmann2018likelihood}
Lueckmann, J., Bassetto, G., Karaletsos, T., and Macke, J.
\newblock Likelihood-free inference with emulator networks. arxiv e-prints.
\newblock \emph{arXiv preprint arXiv:1805.09294}, 2018.

\bibitem[Lueckmann et~al.(2021)Lueckmann, Boelts, Greenberg, Goncalves, and
  Macke]{lueckmann2021benchmarking}
Lueckmann, J.-M., Boelts, J., Greenberg, D., Goncalves, P., and Macke, J.
\newblock Benchmarking simulation-based inference.
\newblock In \emph{International Conference on Artificial Intelligence and
  Statistics}, pp.\  343--351. PMLR, 2021.

\bibitem[Oord et~al.(2018)Oord, Li, and Vinyals]{oord2018representation}
Oord, A. v.~d., Li, Y., and Vinyals, O.
\newblock Representation learning with contrastive predictive coding.
\newblock \emph{arXiv preprint arXiv:1807.03748}, 2018.

\bibitem[Papamakarios \& Murray(2016)Papamakarios and Murray]{Papamakarios2016}
Papamakarios, G. and Murray, I.
\newblock Fast e-free inference of simulation models with {Bayesian}
  conditional density estimation.
\newblock In \emph{Advances in {Neural} {Information} {Processing} {Systems}},
  pp.\  1036--1044, May 2016.
\newblock URL \url{http://arxiv.org/abs/1605.06376}.
\newblock arXiv: 1605.06376 Issue: Nips ISSN: 10495258.

\bibitem[Papamakarios et~al.(2018)Papamakarios, Sterratt, and
  Murray]{Papamakarios2018}
Papamakarios, G., Sterratt, D.~C., and Murray, I.
\newblock Sequential neural likelihood: {Fast} likelihood-free inference with
  autoregressive flows, May 2018.
\newblock URL \url{http://arxiv.org/abs/1805.07226}.
\newblock arXiv: 1805.07226 Publication Title: arXiv.

\bibitem[Papamakarios et~al.(2019)Papamakarios, Nalisnick, Rezende, Mohamed,
  and Lakshminarayanan]{Papamakarios2019}
Papamakarios, G., Nalisnick, E., Rezende, D.~J., Mohamed, S., and
  Lakshminarayanan, B.
\newblock Normalizing {Flows} for {Probabilistic} {Modeling} and {Inference}.
\newblock December 2019.
\newblock URL \url{http://arxiv.org/abs/1912.02762}.
\newblock arXiv: 1912.02762.

\bibitem[Poole et~al.(2019)Poole, Ozair, Van Den~Oord, Alemi, and
  Tucker]{poole2019variational}
Poole, B., Ozair, S., Van Den~Oord, A., Alemi, A., and Tucker, G.
\newblock On variational bounds of mutual information.
\newblock In \emph{International Conference on Machine Learning}, pp.\
  5171--5180. PMLR, 2019.

\bibitem[Sisson et~al.(2018)Sisson, Fan, and Beaumont]{sisson2018overview}
Sisson, S., Fan, Y., and Beaumont, M.
\newblock Overview of approximate bayesian computation. arxiv e-prints, art.
\newblock \emph{arXiv preprint arXiv:1802.09720}, 2018.

\bibitem[Talts et~al.()Talts, Betancourt, Simpson, Vehtari, and
  Gelman]{Talts2018}
Talts, S., Betancourt, M., Simpson, D., Vehtari, A., and Gelman, A.
\newblock Validating bayesian inference algorithms with simulation-based
  calibration.
\newblock \doi{10.48550/arxiv.1804.06788}.
\newblock URL \url{https://arxiv.org/abs/1804.06788v2}.

\end{thebibliography}
